\documentclass[runningheads]{llncs}
\usepackage{multirow}
\usepackage{graphicx}
\usepackage{algorithm2e,setspace}
\usepackage{footnote}
\usepackage{amsmath}
\usepackage{subfigure}
\usepackage[misc,geometry]{ifsym}
\usepackage[colorlinks,linkcolor=black]{hyperref}
\begin{document}
\title{Robust End-to-End Offline Chinese Handwriting Text Page Spotter with Text Kernel}
\titlerunning{Robust Offline Chinese Handwriting Text Page Spotter}
%
\author{Zhihao Wang \inst{1} \and Yanwei Yu  (\Letter)  \inst{1,2} \and Yibo Wang \inst{1} \and Haixu Long \inst{1} \and Fazheng Wang \inst{1}}
%
\authorrunning{Zhihao Wang et al.}
%
\institute{School of Software Engineering,\\University Of Science And Technology Of China\\
\email{ \{cheshire, wrainbow, hxlong, wfzc\}@mail.ustc.edu.cn}\\
\and Suzhou Institute for Advanced Research,\\University Of Science And Technology Of China \\
\email{ywyu@ustc.edu.cn}
}

%
\maketitle              
\begin{abstract}
Offline Chinese handwriting text recognition is a long-standing research topic in the field of pattern recognition.
In previous studies, text detection and recognition are separated, which leads to the fact that text recognition is highly dependent on the detection results.
In this paper, we propose a robust end-to-end Chinese text page spotter framework. 
It unifies text detection and text recognition with text kernel that integrates global text feature information to optimize the recognition from multiple scales, which reduces the dependence of detection and improves the robustness of the system. 
Our method achieves state-of-the-art results on the CASIA-HWDB2.0-2.2 dataset and ICDAR-2013 competition dataset.
Without any language model, the correct rates are 99.12\% and 94.27\% for line-level recognition, and 99.03\% and 94.20\% for page-level recognition, respectively. Code will be available at \href{https://github.com/grsgth/Offline-Chinese-Handwriting-Text-Page-Spotter-with-Text-Kernel}{GitHub}.

\end{abstract}

\section{Introduction}
Offline Chinese handwriting text recognition remains a challenging problem. The main difficulties come from three aspects: the variety of characters, the diverse writing styles, and the problem of character-touching. In recent years, methods based on deep learning have greatly improved recognition performance. There are two types of handwriting text recognition methods, page-level recognition methods and line-level recognition methods. For offline Chinese handwriting text recognition, most studies are based on line-level text recognition.

\begin{figure}
\centering
\subfigure[Recognition of the distorted text image.]{
\centering
\includegraphics[width=0.9\textwidth]{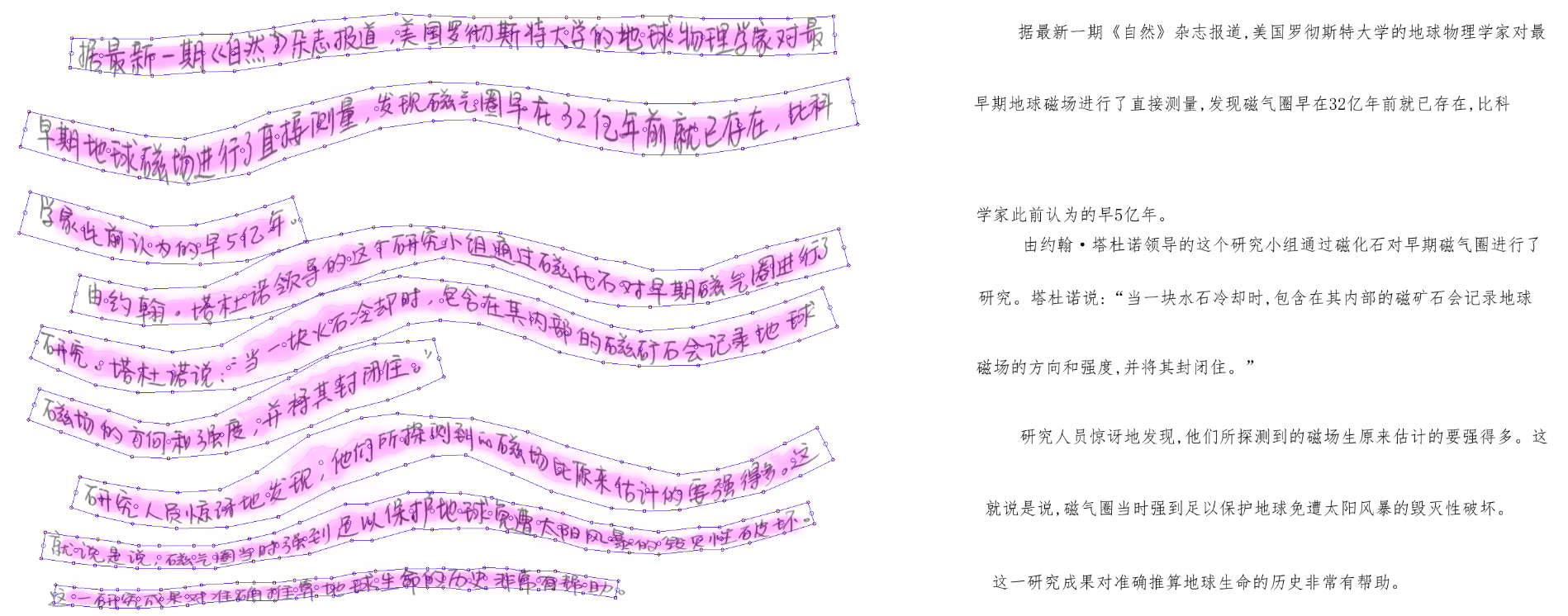}

}
\subfigure[Recognition of the dense text image.]{
\centering
\includegraphics[width=0.9\textwidth]{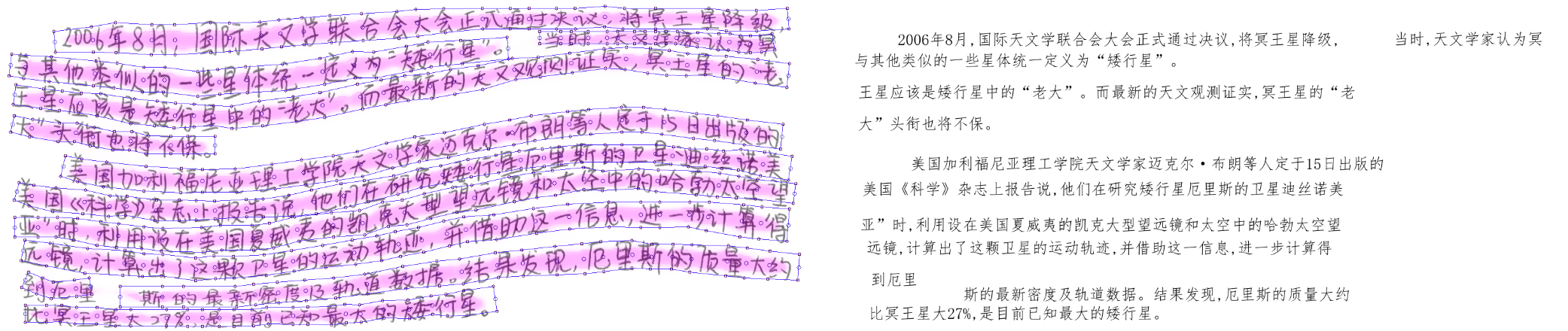}
}
\caption{Results of End-to-end detection and recognition. The pink areas are the segmentation results of kernel areas and the text boxes are generated by the center points generation algorithm.} \label{segreco}
\end{figure}

For line-level recognition, it is mainly classified into two research directions: over-segmentation methods and segmentation-free methods. Over-segmentation methods first over-segment the input text line image into a sequence of primitive segments, then combine the segments to generate candidate character patterns, forming a segmentation candidate lattice, and classify each candidate pattern to assign several candidate character classes to generate a character candidate lattice.
Wang Q F et al.~\cite{WangQF} first proposed the over-segmentation method from the Bayesian decision view and convert the classifier outputs to posterior probabilities via confidence transformation.
Song W et al.~\cite{SongW} proposed a deep network using heterogeneous CNN to obtain hierarchical supervision information from the segmentation candidate lattice.

However, the over-segmentation method has its inherent limitations. If the text lines are not correctly segmented, it brings great difficulties to subsequent recognition. 
The segmentation-free method based on deep learning is proposed to solve this problem.
Messina et al.~\cite{Messina} proposed multidimensional long-short term memory recurrent neural networks (MDLSTM-RNN) using Connectionist Temporal Classifier~\cite{ctc}(CTC) as loss function for end-to-end text line recognition.
Xie et al.~\cite{XieC} proposed a CNN-ResLSTM model with a data preprocessing and augmentation pipeline to rectify the text pictures to optimize recognition.
Xiao et al.~\cite{XiaoS} proposed a deep network with Pixel‑Level Rectification to integrate pixel-level rectification into CNN and RNN-based recognizers.

For page-level recognition, it can be classified into two-stage recognition methods or end-to-end recognition methods. 
The two-stage recognition methods apply two models for text detection and recognition respectively.
End-to-end methods gradually compress the picture into several lines or a whole line of feature maps for recognition.
Bluche et al.~\cite{Bluche} proposed a modification of MDLSTMRNNs to recognize English handwritten paragraphs. 
Yousef et al.~\cite{Yousef} proposed the OrigamiNet which constantly compresses the width of the feature map to 1 dimension for recognition.
The two-stage methods generally detect the text line first and then cut out the text line for recognition. 
Li X et al.~\cite{psenet} proposed segmentation-based PSENet with progressive scale expansion to detect arbitrary-shaped text lines.
Liu Y et al.~\cite{abcnet} proposed ABCNet based on third-order Bezier curve for curved text line detection. 
Liao M et al.~\cite{masktextspotter} proposed Mask TextSpotter v3 with a Segmentation Proposal Network (SPN) and hard RoI masking for robust scene text spotting.

It can be noted that the methods of page-level recognition without detecting text lines lose the information of the text location. If the text layout is complicated, it is difficult to correctly recognize text images with these methods. The method of Liao M et al.~\cite{masktextspotter} applying hard RoI masking into the RoI features instead of transforming text lines may process very big feature maps and loses information when resizing feature maps.
Regardless of the line-level or the two-stage recognition methods, the location information of the text line is obtained first, and the text line is segmented and then recognized, which actually separates the detection and recognition.

We think the detection and recognition should not be separated. 
The detection can only provide the local information of text lines, which makes it difficult to utilize the global text image information during recognition. 
Whether the detection box is larger or smaller than the ground truth box, it causes difficulties for subsequent recognition. 
This is because the alignment of text line images is in the original text page image, and the robustness is not enough for the recognition. 
And the detection of text lines is so important that if the text line cannot be detected well, the text line image cannot be accurately recognized.
We believe that the key to text recognition lies in accurately recognizing text and we just need to know the approximate location of the text, instead of precise detection.

In this paper, we propose a robust end-to-end Chinese text page recognition framework with text kernel segmentation and multi-scale information integration to unify text detection and recognition.
Our main contributions are as follows:
\begin{itemize}
\item [1)]We propose a novel end-to-end text page recognition framework, which utilizes global information to optimize detection and recognition.
\item [2)]
We propose a method to align text lines with the text kernel, which is based on center-lines to extract text line strips from feature maps.
\item [3)]
We propose a text line recognition model with multi-scale information integration, which uses TCN and Self-attention instead of RNNs.
\item [4)]
We have done a series of experiments to verify the effectiveness of our model and compare it with other state-of-the-art methods. We achieve state-of-the-art performance on both the CASIA-HWDB dataset and the ICDAR-2013 dataset. The page-level recognition performance of our method is even better than the line-level recognition methods.
\end{itemize}

\section{Method}
\begin{figure}
\includegraphics[width=\textwidth]{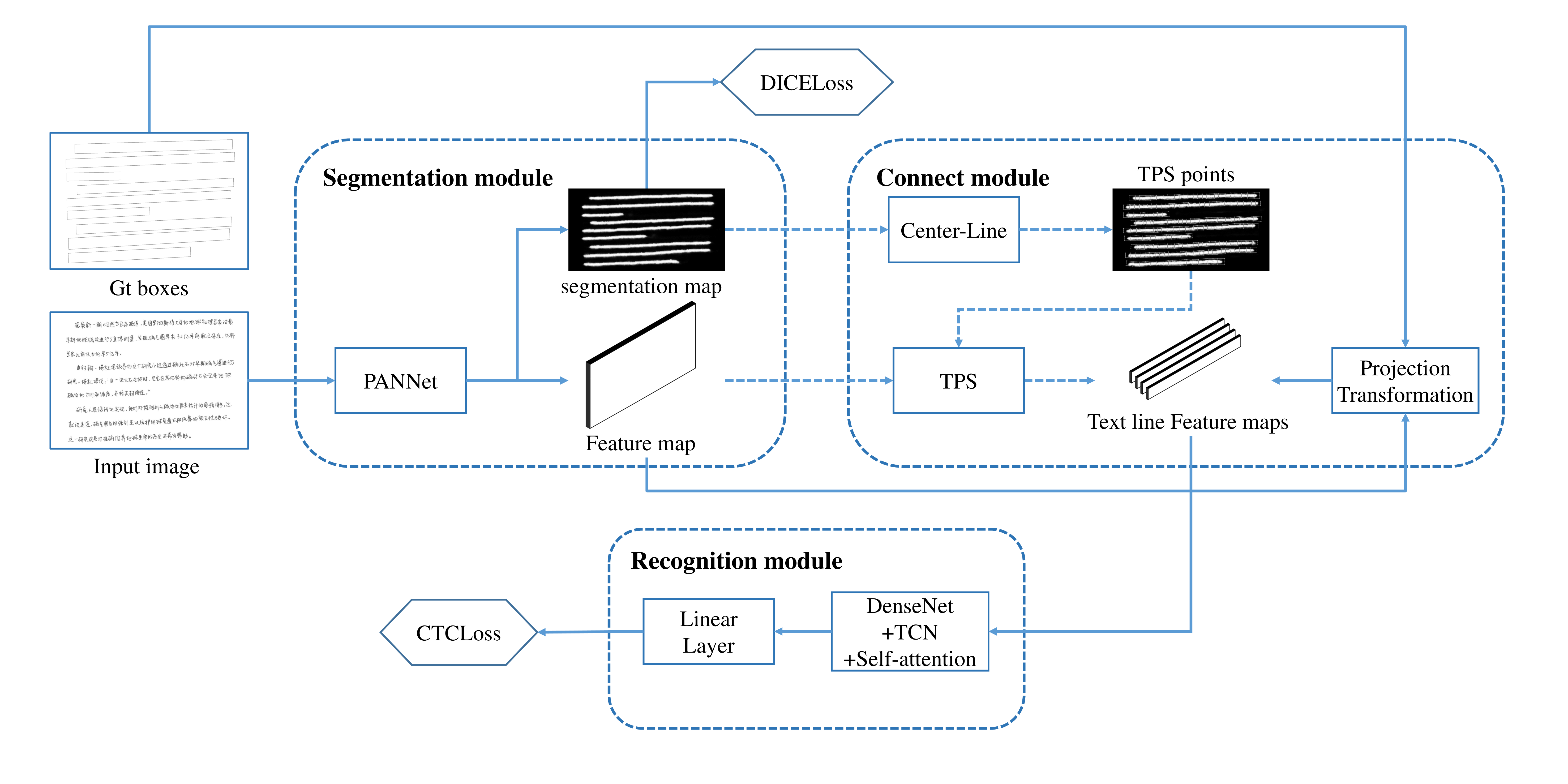}
\caption{Overview of our text page recognition framework. 
The dashed arrow denotes that the text line feature maps are transformed by the center lines with the kernel areas when predicting. } \label{framework}
\end{figure}

The framework of our method is depicted in Figure~\ref{framework}. 
This framework consists of three modules for text detection and recognition.
The segmentation module is used to generate the segmentation map of the kernel area of the text line and the feature map of the text page. 
The connection module is introduced to extract the text line feature map according to the segmentation map. 
The recognition module is used for text line recognition which is based on DenseNet~\cite{densenet} with TCN and Self-attention. 
The segmentation and recognition results are shown in Figure~\ref{segreco}. 

\subsection{Segmentation module}
The segmentation module processes the input image to generate a feature map and a segmentation map that are one-quarter the size of the original image.
In this part, We mainly utilize the network structure of PANNet~\cite{pannet} for its good performance on text segmentation of arbitrary shapes. We use ResNet34~\cite{resnet} as its backbone and change the strides of the 4 feature maps generated by backbone to 4, 4, 8, 8 with respect to the input image. We extract larger feature maps because we need fine-grained feature information for text recognition. We retain the Feature Pyramid Enhancement Module(FPEM) and Feature Fusion Module(FFM) of PANnet for extracting and fusing feature information of different scales and the number of repetitions of the FPEM is 4. The size of the text line kernel area we set is 0.6 of the original size, which is enough to distinguish different text line regions.

\subsection{Connection module}
The connection module is used to extract the text line feature map according to the segmentation map. 
We believe that the feature map contains high-dimensional information than the original image, and the feature map can gather the text information in the kernel area, which makes the extraction of the text line feature map more robust. 
We transform the feature map randomly with the kernel area as the center, such as perspective transformation, so that the text feature information is concentrated in the kernel area.
We scale all the text line feature maps to a height of 32-pixel for subsequent recognition. 

\begin{figure}

\includegraphics[width=\textwidth]{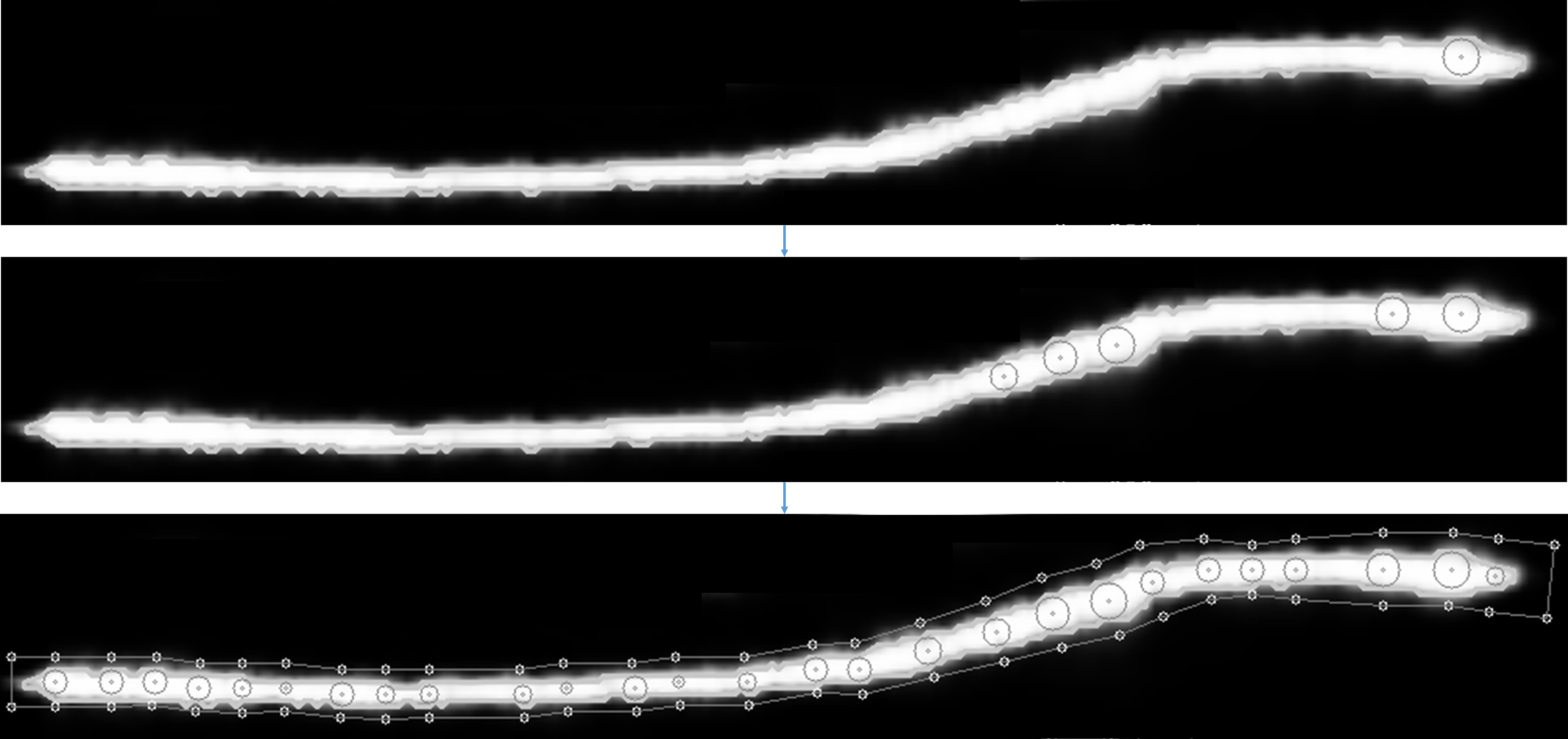}
\caption{The center points are generated by continuously finding the center of the current largest inscribed circle. } 
\label{pointgen}
\end{figure}

We assume that the text line is a strip, and its center-line can be considered as passing through the center of each character. Generally speaking, the length of a text line is greater than its height, and each text line haves one center-line. 
We use the inscribed circle of the text line strip to find the text center-line. We assume that the center of the largest inscribed circle of the text line strip falls on the text center-line.

During training, we use perspective transformation to align the text line according to the ground truth text box. And the aligned text line feature maps are randomly affine transformed in the direction of the interior for data enhancement which allows the model to learn to concentrate the text feature information in the kernel area.

In the evaluation, we align text lines based on the segmentation map. But in this way we can only get the contour of the text line kernel region, we also need to get the trajectory of the text line. Here we propose the Algorithm~\ref{pointalg} to generate the center-line based on the contour. We need to calculate the shortest distance between each inside point and the contour boundary, so we can get the maximum inscribed circle radius of each point in the contour. $Distance$ represents the calculation of the Euclidean distance between two points. $Area$ and $Perimeter$ represent the calculation of the area and perimeter of the contour.

\begin{algorithm}
\caption{Center points generation}
\label{pointalg}
\SetAlgoLined
\KwData{$contour$}
\KwResult{$Points_{center}$}
Calculate the shortest distance between each inside point and the contour boundary as $distances$;\\
$Points_{center}=\left[ \right]$\;
$min_R = Area(contour) / Perimeter(contour)$\;

\While{$Max(distances)>min_R$}
{
    $distance_{max}=Max(distances)$\;
    Delete $distance_{max}$ from $distances$\;
    Add the $point$ with its $distance_{max}$ to $Points_{center}$ \;
    \For{$point_i \in contour$}{
        \eIf{$Distance(point,point_i)<4*min_R$}{
        $distance$ of $point_i$ = 0
    }{
        pass
    }
    }
    

}
\end{algorithm}

This algorithm continuously finds the largest inscribed circle in the contour and adds its center to the point set of the contour center-line and we also add the points that extend at both ends of the center-line to the center point set. Figure~\ref{pointgen} shows the process of generating the center point.

We assume that the abscissa of the start point is smaller than the endpoint. And We can reorder them by the distance between the center points, as shown in Algorithm~\ref{reorder}. 
We use thin-plate-spline(TPS)~\cite{tps} interpolation to align the images, which can match and transform the corresponding points and minimize the bending energy generated by the deformation. 

\begin{algorithm}
\caption{Center points reordering}
\label{reorder}
\SetAlgoLined
\KwData{$Points_c$ of length $l$, $min_R$}
\KwResult{$Points\_reorder_c$}
$Points\_reorder_c=\left[ \right]$\;
\eIf{Len($Points_c$)==1}{
    $Points\_new_c$ =[$Points_c$] \;
    }{
    $Points\_new_c$ = $Points_c\left[:2 \right]$\;
    \For{$i\leftarrow 2$ \KwTo $l$}{
        $left_d = Distance(point_i,Points\_new_c\left[ 0\right])$\;
        $right_d = Distance(point_i,Points\_new_c\left[ -1\right])$\;
        $left\_right_d = Distance(Points\_new_c\left[ 0\right],Points\_new_c\left[ -1\right])$\;
        \eIf{$right_d>left\_right_d$ and $right_d >left_d$}{
            Insert $point_i$ to $Points\_new_c$ in position $0$\;
        }{
        \eIf{$left_d>left\_right_d$ and $right_d <left_d$}{
            Insert $point_i$ to $Points\_new_c$ in position $-1$\;
        }{
            $mid=\arg\min_{j \in [1,l]} (Distance(point_i,Points_c\left[ j\right])+Distance(point_i,Points_c\left[ j-1\right]))$\;
            Insert $point_i$ to $Points\_new_c$ in position $mid$\;
        }
        }

    }
      
    }
\end{algorithm}

\begin{figure}

\includegraphics[width=\textwidth]{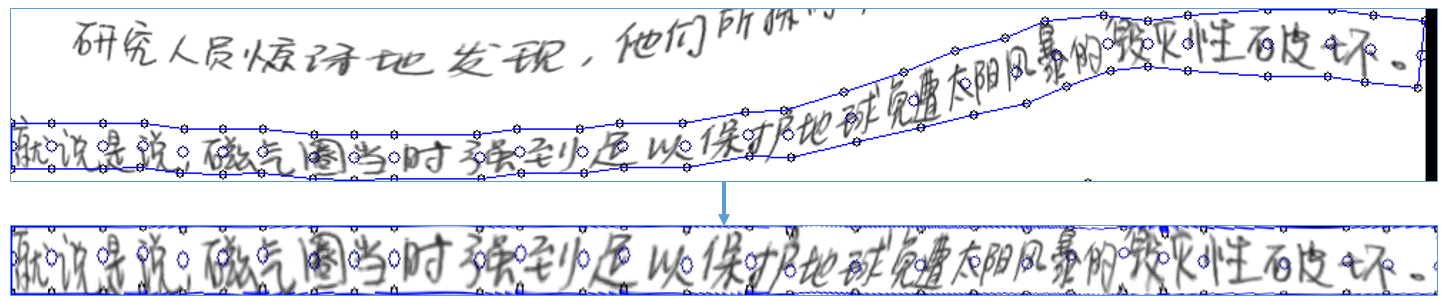}
\caption{TPS transformation on the text picture. Actually, our method performs transformation operations in the feature map.} 
\label{tps}
\end{figure}

According to the points and radius of the center-line, we can generate coordinate points for TPS transformation, as shown in Figure~\ref{tps}. Through TPS transformation, we can transform irregular text line strips into rectangles.

\subsection{Recognition module}

\begin{figure}
\includegraphics[width=\textwidth]{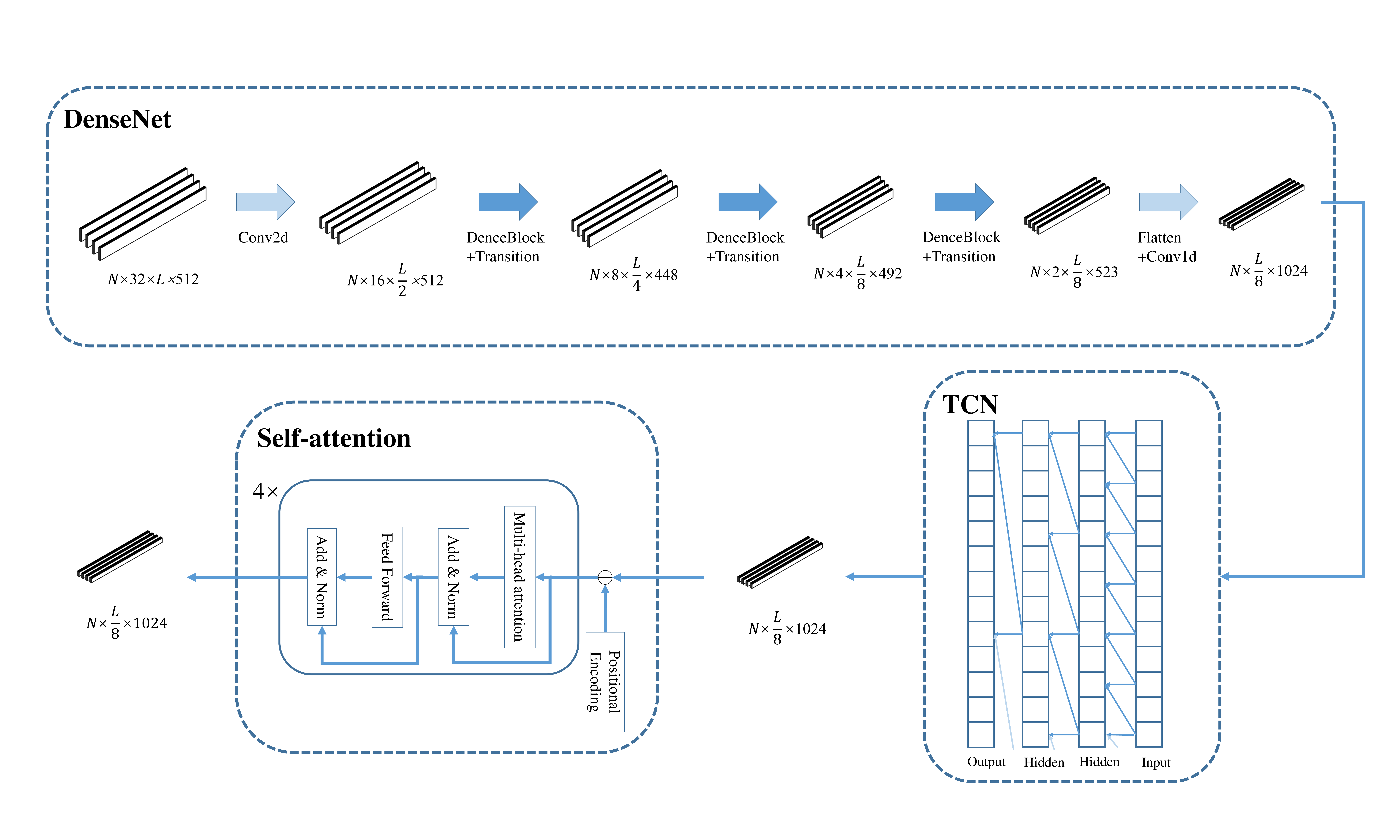}
\caption{Network architecture of recognition module.} \label{recomodel}
\end{figure}

This module is roughly equivalent to traditional text line recognition, except that we replaced the text line image with the text line feature map. Most methods use RNNs to construct semantic relations for time-series feature maps. We think that RNNs have three main disadvantages: 1) it is prone to gradient disappearance; 2) the calculation speed is slow; 3) the effect of processing long text is not good. We designed a multi-scale feature extraction network for text line recognition without using RNNs. 
And compared with CNN, TCN and Self-attention have stronger long-distance information integration capabilities. 
The structure of the recognition module is shown in Figure~\ref{recomodel}

For the task of text line recognition, we think there are three levels of recognition. The first is to extract features of each character based on each character's image information. The second is to perform auxiliary feature extraction based on several characters surrounding every single character. The third is to optimize features based on global text information.

For the first level, we use DenseNet as the backbone to perform further feature extraction on the feature map, gradually compress the height of the feature map to 1, and convert it into a time-sequential feature map. The multiplicative factor for the number of bottleneck layers is 4. There are 32 filters added to each Dense Layer. We added the CBAM~\cite{cbam} layer before each Dense Block to exploit both spatial and channel-wise attention. And the other detailed configuration of DenseNet is presented in Table~\ref{densenet}.

\begin{table}[]
\centering
\caption{The structure of DenseNet.}
\label{densenet}
\begin{tabular}{|c|c|c|}

\hline
moudle           & Output size &    Config                                       \\ \hline

Convolution 1      & 16×$\frac{W}{2}$      & $3 \times 3 \ conv,\ stride \ 2, 512\rightarrow512$ \\ \hline

Dense Block 1      & 16×$\frac{W}{2}$      & $\begin{bmatrix}
    1 \times 1 & conv \\
    3 \times 3 & conv \\
\end{bmatrix} \times 4,\ 512 \rightarrow 640 $                    \\ \hline

Transition Layer 1 & 8×$\frac{W}{4}$       & $3 \times 3 \ conv,\ stride \ 2 \times 2,\ 640\rightarrow448$ \\ \hline
Dense Block 2      & 8×$\frac{W}{4}$       & $\begin{bmatrix}
    1 \times 1 & conv \\
    3 \times 3 & conv \\
\end{bmatrix} \times 8,\ 448\rightarrow704   $                  \\ \hline

Transition Layer 2 & 4×$\frac{W}{8}$      & $3\times3 \ conv,\ stride \ 2 \times 2,\ 704 \rightarrow 492$ \\ \hline

Dense Block 3     & 4×$\frac{W}{8}$       & $\begin{bmatrix}
    1 \times 1 & conv \\
    3 \times 3 & conv \\
\end{bmatrix} \times 8,\ 492 \rightarrow 748    $                 \\ \hline

Transition Layer 3 & 2×$\frac{W}{8}$       & $3 \times 3 \ conv,\ stride \ 2 \times 1,\ 748 \rightarrow 523$ \\ \hline

Flatten          & $\frac{W}{8}$         & $523 \rightarrow 1046 $                   \\ \hline 

Convolution 2      & $\frac{W}{8}$         & $1 \ conv, \ 1046 \rightarrow 1024  $         \\ \hline
\end{tabular}
\end{table}

We use the Temporal Convolutional Network(TCN)~\cite{tcn} for the second level of feature extraction. The dilated convolution is used on TCNs to obtain a larger receptive field than CNN. Compared with RNNs, it performs parallel calculations, and the calculation speed is faster, and the gradient is more stable. A 4-layer TCN is introduced in our model and the maximum dilation is 8 which makes the model has a receptive field of size 32.

For the third level, we adopt the structure of Self-attention for global information extraction. We only use the Encoder layer of Transformer~\cite{transformer}, and the multi-head self-attention mechanism is the core to build global feature connections. The self-attention mechanism is mainly used in the field of natural language processing, but the text obtained by text recognition also has semantic features. We can use Self-attention to construct semantic association to assist recognition. A 4-layer Self-attention encoder is applied in our model, and the hidden layer dimension is 1024 with 16 parallel attention heads.

\subsection{Loss Function}
The loss function is as follows:

\begin{equation}
L = L_{text} + \alpha L_{kernel}
\end{equation}

$L_{text}$ is the character recognition loss, calculated with CTC, and $L_{kernel}$ is the loss of the kernel, and $\alpha$ is to balance the importance of the two, we set $\alpha$ to 0.1.

Considering the imbalance between the text area and the non-text area, we use the dice coefficient as the method to evaluate the segmentation results.

\begin{equation}
L_{kernel}=1-\frac{2\sum_{i} P_{text}(i)G_{text}(i)}{\sum_{i} {P_{text}(i)}^2 + \sum_{i} {G_{text}(i)}^2}
\end{equation}

$P_{text}(i)$ and $G_{text}(i)$ represent the value of the $i$th pixel in the segmentation result and the ground truth of the text regions respectively. The ground truth of the text regions is a binary image, in which the text pixel is 1 and the non-text pixel is 0.

\section{Experiment}
\subsection{Dataset}
We use CASIA-HWDB~\cite{hwdb} as the main dataset, which has been divided into the train set and test set. 
It includes offline handwriting isolated character pictures and offline unconstrained handwritten text page pictures. 
CASIA-HWDB1.0-1.2 contains 3,721,874 isolated character pictures of 7,356 classes. 
CASIA-HWDB2.0-2.2 contains 5,091 text page pictures, containing 1,349,414 characters of 2,703 classes. 
CASIA-HWDB2.0-2.2 is divided into 4,076 training samples and 1,015 test samples. 
ICDAR-2013 Chinese handwriting recognition competition Task 4~\cite{icdar} is a dataset containing 300 test samples, containing 92,654 characters of 1,421 classes. The isolated character pictures of CASIA-HWDB1.0-1.2 are used to synthesize 20,000 text page pictures, containing 9,008,396 characters. The corpus for synthesizing is a dataset containing 2.5 million news articles.

\begin{figure}
\includegraphics[width=\textwidth]{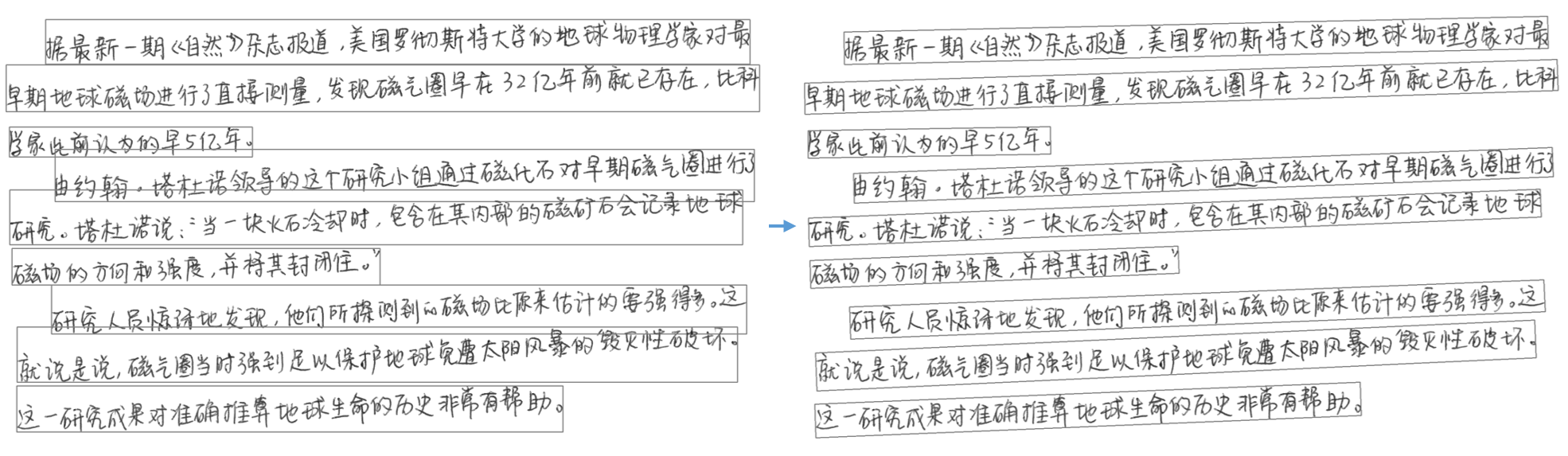}
\caption{Regenerate the text boxes with the smallest enclosing rectangle.} \label{change_box}
\end{figure}

We regenerate the text line boxes by using the smallest enclosing rectangle method according to the outline of the text line, as shown in Fig~\ref{change_box}. 
We use the same method as PANNet to generate the text kernels that adopt the Vatti clipping algorithm~\cite{clipping} to shrink the text regions by clipping d pixels. 
The offset pixels d can be determined as $d = A(1 - r^2)/L$, where $A$ and $L$ are the area and perimeter of the polygon that represents the text region, and $r$ is the shrink ratio, which we set to 0.6.

We convert some confusing Chinese symbols into English symbols, such as quotation marks and commas. 

\subsection{Experimental Settings}
We use PyTorch to implement our system. We train the whole system with the Adam optimizer and the batch size is set to 4. We resize the pictures with a length not greater than 1200 or a width not greater than 1600. No language model is used to optimize the recognition results.
We conducted two main experiments, and the rest of the experimental settings are as follows:

\begin{itemize}
\item [1)]
Train set: train set of CASIA-HWDB2.0-2.2;\\
Test set: test set of CASIA-HWDB2.0-2.2;\\
Learning rate: initialized to $1 \times 10^{-4}$ and multiplied by $0.9$ every 2 epochs;\
Train epoch: 50.\\

\item [2)]
Train set: train set of CASIA-HWDB2.0-2.2 and 20,000 synthesized text page pictures; \\
Test set: test set of CASIA-HWDB2.0-2.2 and test set of ICDAR-2013;\\
Learning rate: initialized to $5 \times 10^{-5}$ and multiplied by $0.9$ every 2 epochs;\\
Train epoch: 50;\\
The model weight of Experiment 1 is used for initialization, except for the final fully connected layer.

\end{itemize}

\begin{algorithm}
\caption{Kernel boxes grouping}
\label{group_box}
\SetAlgoLined
\KwData{$kernel\_boxers$ with length $m$, $gt\_boxes$ with length $l$}
\KwResult{$kernel\_boxers\_group$}
$kernel\_boxers\_group$ is a list with length $l$\;
\For{$i\leftarrow 0$ \KwTo $m$}{
    $index=\arg\max_{j \in [0,l-1]} (IOU(kernel\_boxers[i],gt\_boxes[j])$\;
    add $kernel\_boxers[i]$ to $kernel\_boxers\_group[index]$
}
\end{algorithm}

\subsection{Experimental Results}
\textbf{With ground truth box}

The ground truth boxes are used for segmentation so that our model is equivalent to the text line recognition model. 
But compared with the previous text line recognition model, our model can utilize the global information of the text page for recognition, which has stronger anti-interference and robustness.
The experimental results are shown in Table~\ref{allcompare}.
Notably, our method is superior to the previous results by a large absolute margin of 1.22\% CR and 2.14\% CR on CASIA-HWDB2.0-2.2 and ICDAR-2013 dataset without the language model. 
The recognition performance of our model in CASIA-HWDB2.0-2.2 dataset is even better than the method using the language model.

\begin{figure}
\includegraphics[width=\textwidth]{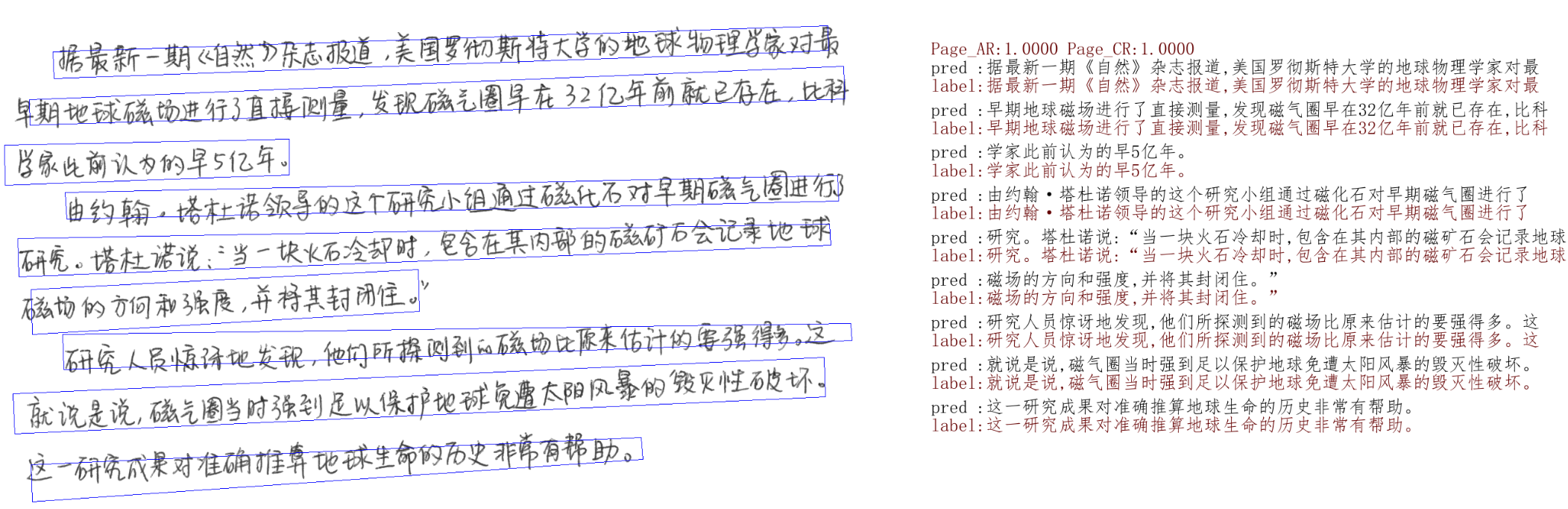}
\caption{Recognize text picture when the text box is not right.} 
\label{gt_box}
\end{figure}

Even when the text boxes do not include the whole text line, our model can recognize the text line correctly, as shown in Figure~\ref{gt_box}.
We also conduct an experiment with the size of the gt boxes randomly changed that the outline points of gt boxes are randomly moved by -0.2 to 0.2 of the width in the vertical direction and moved by -1 to 1 of the width in the horizontal direction, and the CR only dropped by about 0.5\%, as shown in Table~\ref{allcompare}, which shows that our model is very robust.
Because we perform segmentation and transformation of text lines in the feature map of the text page, we expand or shrink the transformation box so that the text information is concentrated in the core area of the text. Low-dimensional information is more sensitive, if such data enhancement is performed directly on the original picture, some text will become incomplete and cannot be correctly recognized. 
\\

\begin{table}[]
\centering
\caption{Comparison with the start-of-the-art methods.}
\label{allcompare}
\begin{tabular}{|c|c|cc|cc|cc|cc|}
\hline
\multicolumn{2}{|c|}{\multirow{3}{*}{Method}}   & \multicolumn{4}{c|}{Without LM}                             & \multicolumn{4}{c|}{With LM}                                                \\ \cline{3-10} 
\multicolumn{2}{|c|}{}                          & \multicolumn{2}{c|}{HWDB2} & \multicolumn{2}{c|}{ICDAR-2013} & \multicolumn{2}{c|}{HWDB2}           & \multicolumn{2}{c|}{ICDAR-2013}       \\ \cline{3-10} 
\multicolumn{2}{|c|}{}                          & CR(\%)            & AR(\%)      & CR(\%)         & AR(\%)        & CR(\%) & AR(\%) & CR(\%) & AR(\%) \\ \hline
\multicolumn{2}{|c|}{Wu~\cite{Wu}}              & -                 & -           & -              & -             & 95.88                       & 95.88  & 96.20                       & 96.20  \\
\multicolumn{2}{|c|}{Peng~\cite{Peng}}          & -                 & -           & 89.61          & 90.52         & -                           & -      & 94.88                       & 94.88  \\
\multicolumn{2}{|c|}{Song~\cite{Song}}          & 93.24             & 92.04       & 90.67          & 88.79         & 96.28                       & 95.21  & 95.53                       & 94.02  \\
\multicolumn{2}{|c|}{Xiao~\cite{XiaoS}}         & 97.90             & 97.31       & -              & -             & -                           & -      & -                           & -      \\
\multicolumn{2}{|c|}{Xie~\cite{XieC}}           & 95.37             & 95.37       & 92.13          & 91.55         & 97.28                       & 96.97  & 96.99                       & 96.72  \\ \hline
\multicolumn{2}{|c|}{Ours, with gt box$^1$}      & \textbf{99.12}    & \textbf{98.84} & -         & -             & -             & -      & -                           & -      \\
\multicolumn{2}{|c|}{Ours, with center-line$^1$} & 99.03             & 98.64       & -              & -            & -                           & -      & -                           & -      \\ 
\multicolumn{2}{|c|}{Ours, with changed gt box$^c$} & 98.58             & 98.22       & -          & -         & -                           & -      & -                           & -      \\\hline
\multicolumn{2}{|c|}{Ours, line level$^l$} & 98.38             & 98.22       & -          & -         & -                           & -      & -                           & -      \\\hline
\multicolumn{2}{|c|}{Ours, with gt box$^2$}      & 98.55             & 98.21        & \textbf{94.27} & \textbf{93.88}         & -             & -      & -                           & -      \\

\multicolumn{2}{|c|}{Ours, with center-line$^2$} & 98.38             & 97.81       & 94.20          & 93.67         & -                           & -      & -                           & -      \\
\hline
\end{tabular}
\\
\footnotesize{$^1$ Experiment 1. $^2$ Experiment 2. \\$^c$ Experiment 1 with the size of the gt boxes randomly changed.\\$^l$Line level recognition with modified recognition module.}\\
\end{table}

\noindent \textbf{With center-line}

It can be noted that the CR calculating with center-line segmentation only drops by about 0.1\%, which fully proves the effectiveness of center-line segmentation.

Since one line of text may be divided into multiple text lines during segmentation, it is necessary to correspond the text box obtained by the segmentation with the ground truth box. We use the method of calculating the intersection over union(IOU) for the match, as shown in Algorithm \ref{group_box}. Before performing this algorithm, it is necessary to sort all the divided boxes according to the abscissa, so that multiple text lines are added to the corresponding group in a left-to-right manner. Each group corresponds to one text line label. If there are extra boxes, that is, all IOUs are calculated as 0, they can be added to any group, which does not affect the calculation of CR and AR. 

After all the segmentation boxes are divided into groups, the detection result is considered correct when the length of the recognition result of the group is greater than or equal to 90\% of the label length. Table~\ref{F1} shows the detection performance of our model.

\noindent \textbf{Effectiveness of the segmentation module}

We can slightly change the structure of the recognition module to make it a text line recognition model. 
The main modification is to change the 3-layer Dense Block to the 5-layer Dense Block, from [4,8,8] to [1,4,8,8,8].
The input data is changed to a 64-pixel text line image with a segmented height and batch size is set to 8, and the other settings are similar to experiment 1.
The recognition results show that the segmentation module can make good use of the global text information to optimize the text feature extraction, thereby improving the recognition performance.
\\

\begin{table}[]
\centering
\caption{Text detection results on CASIA-HWDB2.0-2.2 and ICDAR-2013 dataset.}
\label{F1}
\begin{tabular}{|c|c|c|c|}
\hline
          & Precision(\%) & Recall(\%) & F-measure(\%) \\ \hline
HWDB2     & 99.97    & 99.79 & 99.88    \\
ICDAR-2013 & 99.91    & 99.74 & 99.83    \\ \hline
\end{tabular}
\end{table}

\noindent \textbf{Effectiveness of TCN and Self-attention}

We conduct a series of ablation experiments on CASIA-HWDB2.0-2.2 dataset without data augmentation to verify the effectiveness of TCN and Self-attention, with results shown in Table~\ref{hwdb2compare}. On the test set, 1.08\% and 1.66\%  CR improvements are obtained by the TCN module and the Self-attention module, respectively. And CR is increased by 1.96\% when the TCN module and Self-attention module are 
applied together.

\begin{table}[]
\centering
\caption{Ablation experiment results on CASIA-HWDB2.0-2.2.}
\label{hwdb2compare}
\begin{tabular}{|l|cc|cc|}
\hline
\multicolumn{1}{|c|}{\multirow{2}{*}{Method}} & \multicolumn{2}{c|}{with gt box}     & \multicolumn{2}{c|}{with center-line} \\ \cline{2-5} 
\multicolumn{1}{|c|}{}                        & \multicolumn{1}{c|}{CR(\%)} & AR(\%) & \multicolumn{1}{c|}{CR(\%)}  & AR(\%) \\ \hline
baseline                                      & 97.18                       & 96.81  & 97.09                        & 96.63  \\
+Self-attention                               & 98.74                       & 98.42  & 98.66                        & 98.28  \\
+TCN                                          & 98.26                       & 97.97  & 98.03                        & 97.62  \\
+TCN and Self-attention                       & 99.12                       & 98.84  & 99.03                        & 98.64  \\ \hline

\end{tabular}

\end{table}

\section{Conclusion}
In this paper, we propose a novel robust end-to-end Chinese text page spotter framework. It utilizes global information to concentrate the text features in the kernel areas, which allows the model only need to roughly detect the text area for recognition. TPS transformation is used to align the text lines with center points. TCN and Self-attention are introduced into the recognition module for multi-scale text information extraction. It can perform end-to-end text detection and recognition, or optimize recognition when ground truth text boxes are provided. This architecture can be easily modified that the segmentation module and recognition module can be replaced by better models. Experimental results on CASIA-HWDB2.0-2.2 and ICDAR-2013 datasets show that our method achieves state-of-the-art recognition performance. Future work will be to investigate the performance of our method on the English dataset.

%
%
%
%

\end{document}